\documentclass[sigconf]{acmart}

\usepackage{amsmath,amsfonts}

\usepackage{graphicx}
\usepackage{booktabs}
\usepackage{tabularx}
\usepackage{multirow}

\usepackage{algorithm}
\usepackage{algorithmic}

\usepackage{textcomp}
\usepackage{xcolor}
\usepackage{listings}
\usepackage{cancel}

\usepackage{needspace}
\setcopyright{none}
\renewcommand\footnotetextcopyrightpermission[1]{}

\lstdefinestyle{war-prompt}{
  language={},
  basicstyle=\ttfamily\footnotesize,
  columns=fullflexible,
  keepspaces=true,
  escapeinside={(*@}{@*)},
  showspaces=false,
  showtabs=false,
  showstringspaces=false,
  numbers=none,
  upquote=true,
  tabsize=2,
  breaklines=true,
  breakatwhitespace=true,
  breakautoindent=true,
  breakindent=1em,
  linewidth=\linewidth,
  frame=none,
  framerule=0.25pt,
  framesep=4pt,
  framexleftmargin=0pt,
  framexrightmargin=0pt,
  framextopmargin=0pt,
  framexbottommargin=0pt,
  xleftmargin=0pt,
  xrightmargin=0pt,
  aboveskip=3pt,
  belowskip=5pt
}

\lstdefinestyle{war-input}{
  style=war-prompt,
  basicstyle=\ttfamily\footnotesize,
  frame=single,
  xleftmargin=5pt,
  xrightmargin=5pt,
  aboveskip=5pt,
  belowskip=5pt
}

\newsavebox{\warinputbox}

\newcommand{\toolft}{Tool-Embed-4B\textsuperscript{FT}}

\begin{document}

\title{WAR-CRAFT: Refining Web API Recommendations through Requirement Auditing}

\author{Zishuo Xu}
\affiliation{%
  \institution{University of Illinois Urbana-Champaign}
  \city{Urbana}
  \country{United States}
}
\email{zishuo3@illinois.edu}

\author{Dezhong Yao}
\affiliation{%
  \institution{Huazhong University of Science and Technology}
  \city{Wuhan}
  \country{China}
}
\email{dyao@hust.edu.cn}

\author{Zhiqiang Xu}
\affiliation{%
  \institution{Mohamed bin Zayed University of Artificial Intelligence}
  \city{Abu Dhabi}
  \country{United Arab Emirates}
}
\email{zhiqiang.xu@mbzuai.ac.ae}


\begin{abstract}
Web API recommenders return a set of APIs for a mashup request but do not check it against provider documentation, the main public evidence of an API's capabilities. The set can therefore contain unsupported or redundant APIs and miss requested capabilities. Filling every apparent gap does not fix this, because a gap may already be served, may not need its own API, or may have no documented provider. 
We study post-recommendation refinement, which revises a frozen recommender's output using only the request, the catalog, and provider documentation. 
We propose WAR-CRAFT, which treats refinement as three decisions.
An LLM removes unsupported or subsumed APIs by judging candidates jointly, then audits the remaining set against a requirement checklist derived from the request. 
For each uncovered requirement, a fine-tuned catalog-constrained Resolver proposes a catalog API or abstains, and an independent documentation check admits only supported proposals.
On ProgrammableWeb with 7 heterogeneous upstream recommenders, WAR-CRAFT raises mean SetF1 from 0.558 to 0.644, above the strongest of 5 refinement baselines (62.5\%), and roughly halves functional redundancy (0.127 to 0.065) without changing mean functional coverage. 
Gains range from +2.2 points on generative recommenders to +18.2 on fixed-cutoff rankers. At matched set sizes, WAR-CRAFT still outperforms truncating the upstream output, so the gain is not explained by set-size calibration alone.   Our code is available at \url{https://github.com/ZsXu-enrico/WAR-CRAFT}.
\end{abstract}
\keywords{Web API recommendation, service retrieval, recommendation refinement, functional requirements}
\maketitle

\section{Introduction}
\label{sec:introduction}

Web platforms expose data and functionality through public Web APIs, and Web applications compose these APIs instead of reimplementing each function~\cite{qi2020data,alhosaini2024api}. Such an application,
commonly called a \emph{mashup}~\cite{gong2022dawar}, 
usually has several functional requirements served by different
providers, so selecting its APIs is a \emph{set-level} decision. Two
properties of the Web API ecosystem make this decision hard. Providers
are independent, so an API's implementation is opaque and its
documentation is the main public evidence of what it can do. Providers
also overlap in function, while each added API brings its own
documentation, credentials, and usage terms, so an irrelevant or
duplicated API costs integration effort without adding functionality. LLM-based agents that select Web services or tools for a task face the
same set-level decision~\cite{qin2024toolllm,lu2026tools}.

Web API recommenders estimate the relevance of each API to a mashup
description, using content-based \cite{qi2021compatibility,xu2026warbert},
collaborative-filtering \cite{kang2020nafm,gong2022dawar}, hybrid
\cite{gao2026cvh,pang2026leveraging}, or LLM-based methods
\cite{yang2026lmsr,xu2026ranking,liu2026mars}. 
While they differ in how they score APIs, they share one output form: a ranked list cut at a fixed size or a generated set, neither checked against provider documentation as a set. 
Two gaps remain between recommending APIs and
establishing their suitability for an application. \textbf{1) Unverified set-level suitability.}
Selecting relevant APIs does not by itself establish whether each API
has a documented contribution, whether contributions overlap
unnecessarily, or whether required functions remain uncovered.
Under DOEM's functional labels \cite{rong2026doem}, only 53.25\% of
the functions provided by the initial sets of seven upstream
recommenders in our evaluation are required by the mashup on average, and 12.65\% of
their functional provision is duplicated.
\textbf{2) Missing or inconsistent API-level justification.}
Many existing recommenders return API names or rankings without
explaining how each API serves the request.
Even WAR-R1, which generates rationales, shows mismatches in
our evaluation: 8.56\% of recommended API instances lack
rationales, while 7.53\% of API instances mentioned in its
rationales are absent from the corresponding recommendation sets.
It also shows rationales that misstate
an API's functionality~\cite{xu2026ranking}.
This motivates pairing each selected API with a request-specific
justification checked against the request and API documentation.

\begin{figure*}[t]
  \centering
  \includegraphics[width=0.95\textwidth]{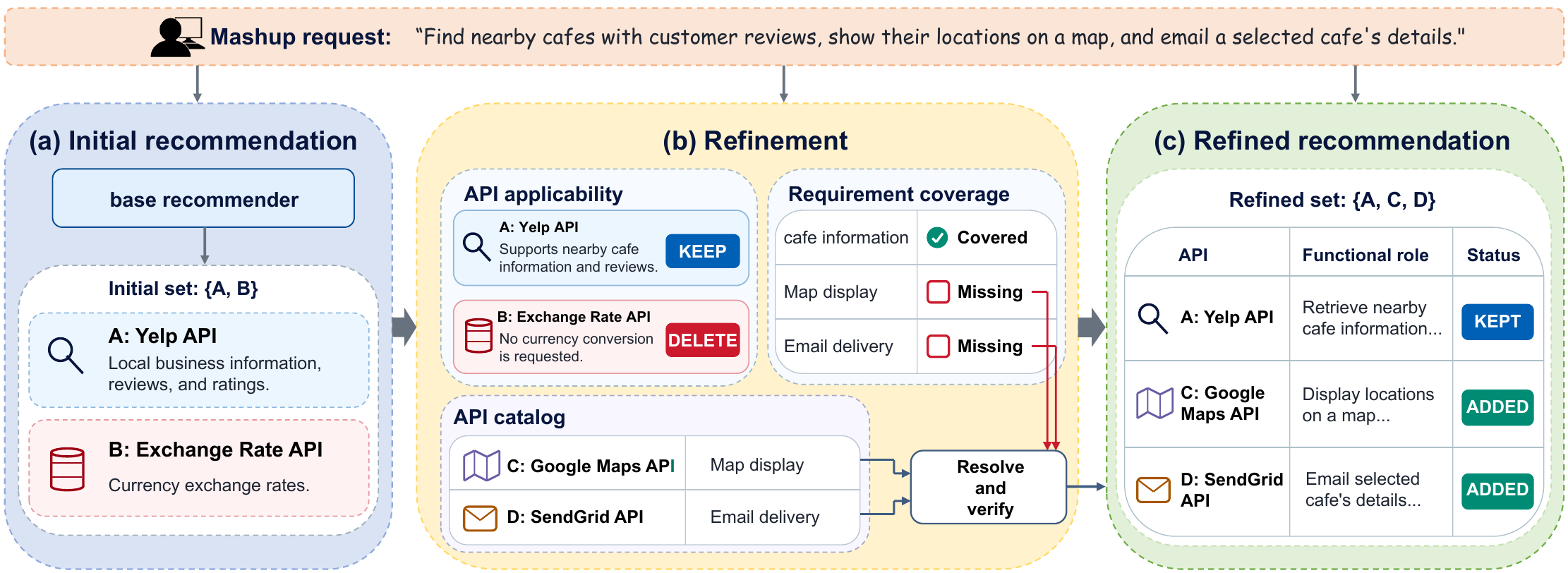}
  \vspace{-0.8em}
  \caption{Post-recommendation refinement for a mashup request. The
  unsupported API B is removed, and the uncovered map-display and
  email-delivery requirements are filled with catalog APIs C and D
  after verification against their documentation. Each selected API
  receives a functional role describing its contribution.}
  \label{fig:refinement-motivation}
  \vspace{-0.8em}
\end{figure*}


Both gaps can be addressed after recommendation by checking the returned set against the request and API documentation without access to the recommender's internals. Doing so, however, does not mean adding an API for every requirement that appears uncovered. A requirement may already be served by a retained API, may not warrant a separate API, or may have no documented provider in the catalog. In our experiments, always adding the top retrieved API for each uncovered requirement lowers SetF1 below filtering alone (0.581 versus 0.591), and the Resolver abstains on 69.7\% of requirement-level decisions. Whether to add is therefore a decision in its own right.

We therefore study \emph{post-recommendation refinement} of Web API sets (Figure~\ref{fig:refinement-motivation}): given a request, an API catalog, and the output of an existing recommender, revise the set by removing unsupported or redundant APIs and selectively adding APIs for unmet requirements, without accessing or retraining the recommender. The refinement is \emph{evidence-constrained}: each decision to keep, remove, or add an API must be justified by the request and API documentation.

We propose \textsc{WAR-CRAFT}, a recommender-agnostic framework that treats refinement as three decisions:  delete, detect gaps, and add or abstain. 
A Role Writer jointly evaluates the initial candidates, removes unsupported or fully subsumed APIs, and assigns functional roles to retained APIs. A Joint Requirement Auditor constructs a requirement checklist and assesses its coverage in a
single step, grounding requirements in the request and coverage in
API documentation. For each uncovered requirement, a fine-tuned,
catalog-constrained Resolver proposes an eligible API or abstains.
A Commit Verifier checks the proposed capabilities against API
documentation and writes a functional role for each accepted
addition. A structured audit trace records the decisions and their supporting evidence. 
We evaluate \textsc{WAR-CRAFT} on ProgrammableWeb with seven heterogeneous upstream recommenders using the same refinement
configuration. It improves SetF1 and CovF1 for all seven sources and reduces average functional redundancy.
It also outperforms truncating the upstream output to its own set size or to the recorded set size, so the gains are not
explained by set size alone, and it exceeds the strongest independent refinement baseline by 1.92 points.
LLM-based evaluation, supplemented by human validation, further indicates that the generated roles are relevant to mashup requirements, grounded in API documentation, and clearly expressed. Our contributions are:
\begin{itemize}
  \item We formulate \emph{post-recommendation refinement} of Web API sets and show that whether to add an API is a decision in its own right: always filling uncovered requirements lowers
SetF1 below filtering alone.

  \item We design \textsc{WAR-CRAFT} around a Resolver that, given the current set, decides for each uncovered requirement whether to add a catalog API or abstain, with documentation-grounded filtering, auditing, verification, and functional roles.

  \item On seven upstream recommenders, WAR-CRAFT improves SetF1 and CovF1 for every source and outperforms set-size truncation, five refinement baselines, and a Toolken+-style abstaining retriever; human audits support these results.
\end{itemize}

\section{Related Work}
\label{sec:related-work}
\subsection{Web API Recommendation}
\label{sec:rw-api}

Web API recommendation includes content-based, collaborative-filtering
(CF), hybrid, and LLM-based approaches. Content-based
methods match mashup and API descriptions through
compatibility-aware text mining~\cite{qi2021compatibility} or hierarchical BERT-based
matching~\cite{xu2026warbert}. CF methods exploit historical
mashup--API interactions~\cite{li2025tssgcf}. Hybrid methods combine
interactions with side information through multi-view
hypergraphs~\cite{gao2026cvh} or causality-inspired context
adjustment~\cite{pang2026leveraging}. 
Complementary work studies API set and bundle
recommendation, considering joint API selection,
compositional matching, and diversity within
recommended sets~\cite{gong2022dawar,
gu2021csbr,yan2025optimal}. STAR sequentially recommends services while updating
residual requirements~\cite{rong2026star}.
 Recent LLM-based methods generate
recommendations or enhance
representations~\cite{qin2025harnessing,yang2026lmsr,sun2026lacl};
WAR-R1 generates API-level rationales~\cite{xu2026ranking}, and MARS
uses multi-agent reasoning~\cite{liu2026mars}. 
These methods primarily produce API recommendations;
\textsc{WAR-CRAFT} instead audits and revises their
frozen outputs against mashup requirements and API
documentation, without modifying the upstream models.

\subsection{Tool Retrieval and Selection}
\label{sec:rw-tool}

Tool retrieval identifies relevant APIs for LLM agents through
learned retrievers~\cite{qin2024toolllm},
reranking~\cite{zheng2024toolrerank}, and document expansion
or intent-based retrieval~\cite{lu2026tools,chen2024re}.
ToolkenGPT learns tool-token embeddings~\cite{hao2023toolkengpt},
while ToolGen uses dedicated tokens for tool retrieval and
calling~\cite{wang2025toolgen}.
Toolken+ incorporates tool documentation into reranking and
introduces a learned reject option~\cite{yakovlev2024toolken+}.
COLT models tool collaboration to improve retrieval
completeness~\cite{qu2024towards}, while AnyTool and MCP-Zero
support tool discovery during task solving~\cite{du2024anytool,fei2025mcp}.
PTR retrieves initial tool bundles, maps their functional
coverage, removes tools that serve no requirement, and
adds tools for unmet requirements~\cite{gao2025task}.
\textsc{WAR-CRAFT} incorporates token-based selection and
abstention into evidence-constrained refinement of frozen
Web API recommendations for mashup development.
Its joint audit identifies which requested capabilities
require external services and checks their coverage by
the retained APIs. Proposed additions undergo a separate
check of their documented support for the requirements
that motivated them.

\subsection{LLM-Based Refinement and Verification}
\label{sec:rw-llm-refinement}
 
LLMs have been used for zero-shot
ranking~\cite{hou2024large} and reranking
supplied candidates, as in
LLM4Rerank~\cite{gao2025llm4rerank}.
Other methods leverage LLM-derived preference
and item-relation knowledge to improve
recommendations~\cite{wang2025unleashing,yang2026agentdr}. Beyond recommendation, LLMs can revise their
outputs through feedback and
reflection~\cite{madaan2023self,shinn2023reflexion}.
RARR further demonstrates evidence-based revision
of unsupported claims~\cite{gao2023rarr}. For fine-grained
verification, FActScore and RefChecker check factual support at the
level of atomic facts and claim triplets, respectively
\cite{min2023factscore,hu2024knowledge}.
Beyond reranking a fixed candidate list,
\textsc{WAR-CRAFT} can remove existing APIs and
select additions from the catalog.
Each proposed addition is accepted only if its documentation
supports the corresponding requirement.

\section{Problem Definition}
\label{sec:problem}

Let $m_i$ be a mashup described by a natural-language request $T_i$, and
let $\mathcal{A}=\{a_1,\ldots,a_N\}$ denote the Web API catalog. Each API
$a\in\mathcal{A}$ has a catalog description $D_a$. A frozen recommender
$P$ maps the mashup request and catalog to an initial API set
\begin{equation}
  S_i^0=P(T_i,\mathcal{A})\subseteq\mathcal{A}.
  \label{eq:initial-set}
\end{equation}

We study \emph{post-recommendation refinement}. The refiner is given only
$(T_i,\mathcal{A},S_i^0)$, with $S_i^0$ kept
in its upstream output order; it neither modifies $P$ nor accesses its
parameters, scores, or intermediate representations. It returns
\begin{equation}
  F(T_i,\mathcal{A},S_i^0)
  =\bigl(\widehat{S}_i,\eta_i,\tau_i\bigr).
  \label{eq:refinement-task}
\end{equation}
Here, $\widehat{S}_i$ is the refined API set. For each
$a\in\widehat{S}_i$, we represent its functional role by
$\eta_i(a)$, a concise natural-language description of one
observable contribution to the mashup supported by $T_i$
and $D_a$. The audit trace $\tau_i$ records refinement
decisions together with their supporting evidence and rationales.

Let $\Delta_i^-$ and $\Delta_i^+$ denote the deleted and added API sets,
respectively. The refined set is
\begin{equation}
\begin{gathered}
  \widehat{S}_i
  =\left(S_i^0\setminus\Delta_i^-\right)\cup\Delta_i^+,\\
  \Delta_i^-\subseteq S_i^0,
  \qquad
  \Delta_i^+\subseteq\mathcal{A}\setminus S_i^0.
\end{gathered}
  \label{eq:refined-set}
\end{equation}

The task imposes three constraints. First, every retained or added API
must have a documented capability that can support a function requested
by $T_i$. Second, a supported initial API may be removed as redundant
only when its documented mashup-relevant contribution is fully subsumed
by another retained initial API, without losing a distinction required
by the mashup. Third, additions must target request-grounded capabilities
not already covered by the retained set. Consequently, the output
cardinality can change through deletion and coverage-driven addition,
rather than being constrained by a fixed cutoff.

\section{WAR-CRAFT}
\label{sec:pipeline}

\begin{figure*}[t]
  \centering
  \includegraphics[width=0.95\textwidth]{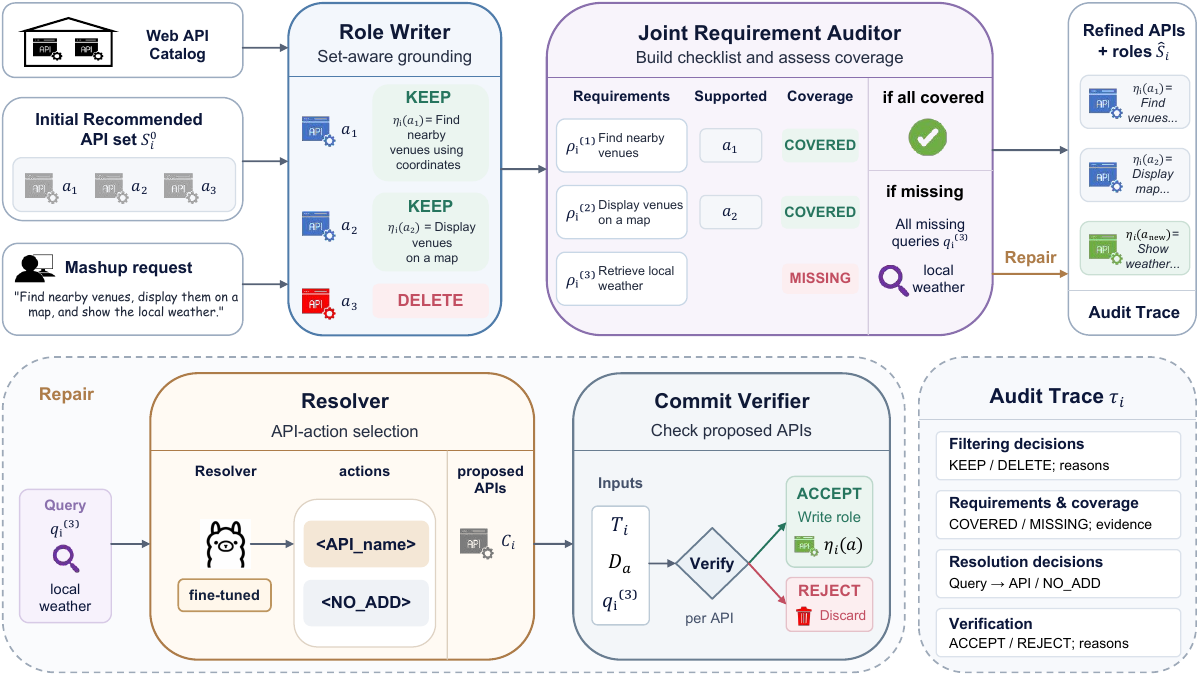}
  \vspace{-0.8em}
  \caption{Overview of \textsc{WAR-CRAFT}. The framework refines
  API recommendations through role writing and joint requirement
  auditing, invoking the Resolver and Commit Verifier only when
  requirements remain uncovered.}
  \label{fig:pipeline}
\end{figure*}

Figure~\ref{fig:pipeline} presents \textsc{WAR-CRAFT}, a recommender-agnostic
framework for refining the output of a frozen Web API recommender.
The Role Writer filters $S_i^0$ into $S_i^{\mathrm{keep}}$
and assigns functional roles. The Joint Requirement Auditor
jointly constructs a checklist from $T_i$ and assesses coverage,
yielding missing-item records $\mathcal{M}_i$.
For each missing query, the Resolver selects an eligible
catalog API or abstains; deduplicated API proposals form
$\mathcal{C}_i$. 
The Commit Verifier checks their documented
support for the originating queries and assigns
functional roles to accepted additions, which
are then merged with $S_i^{\mathrm{keep}}$
to form $\widehat{S}_i$. If the Auditor returns \texttt{COMPLETE}, the Resolver and
Commit Verifier are skipped and
$\widehat{S}_i=S_i^{\mathrm{keep}}$.
The framework returns $(\widehat{S}_i,\eta_i,\tau_i)$
(Section~\ref{sec:problem}); only the Resolver is fine-tuned.

\subsection{Set-Aware API Filtering and Role Writing}
\label{sec:method-roles}

To filter unsupported and functionally redundant APIs, the Role
Writer jointly evaluates all candidates in $S_i^0$ using the
mashup request $T_i$ and their full catalog descriptions in a
single LLM call, allowing redundancy to be judged relative to
other candidates. Canonical API names are withheld to focus
evaluation on documented capabilities. Each \texttt{KEEP}
decision includes a concise, verb-led functional role $\eta_i(a)$,
while each \texttt{DELETE} decision includes a reason code
(\texttt{UNSUPPORTED} or \texttt{REDUNDANT}) and a brief rationale.

The Role Writer first removes candidates whose documented
capabilities do not support any observable function requested
by the mashup. A supported candidate is removed as redundant
only when another retained API fully subsumes its documented
mashup-relevant contribution without losing distinctions
required by the request, such as the data source or
communication channel. Partial overlap is insufficient:
in Best Dining, Google Maps and Yelp Fusion both support
restaurant search but additionally provide map visualization
and restaurant ratings and reviews, respectively.
Among interchangeable candidates, the Role Writer retains
the API whose documented capabilities most directly match
the request.

Candidates assigned \texttt{KEEP} form the retained set
\begin{equation}
  S_i^{\mathrm{keep}}=S_i^0\setminus\Delta_i^-,
  \label{eq:retained-set}
\end{equation}
where $\Delta_i^-$ contains the candidates assigned
\texttt{DELETE}. The retained set, which may be empty,
is passed to the Joint Requirement Auditor for coverage
assessment. Deleted candidates cannot be added later.

\subsection{Joint Requirement and Coverage Audit}
\label{sec:method-joint-audit}

The Joint Requirement Auditor jointly identifies external-service
requirements and checks their coverage by the retained APIs.
In one LLM call, it constructs an ordered checklist
$\mathcal{R}_i$ from $T_i$ and the catalog descriptions
and functional roles of the APIs in $S_i^{\mathrm{keep}}$.
Retained APIs use arbitrary identifiers such as \texttt{A1},
with canonical names and recommendation ranks withheld.
Inspired by candidate-based checklist generation for response
evaluation~\cite{viswanathan2026checklists}, the Auditor uses
retained API functions to guide decomposition and clarify
external-service boundaries. Each requirement must be justified
by $T_i$, whether or not the retained APIs support it.
Coverage judgments rely on API descriptions; roles provide
context but cannot establish undocumented capabilities.
The checklist covers external-service capabilities explicitly
requested or unambiguously required by the stated workflow, excluding
optional enrichments and generic infrastructure unless requested. Each record contains a distinct, atomic, verb-led requirement $\rho_i^{(j)}$ 
and a verbatim span $e_i^{(j)}$ from $T_i$ stating the need or the workflow step that requires it.

A requirement is marked \texttt{COVERED} only if at least
one retained API description directly supports its entire
atomic capability in the mashup context. The Auditor records
all such APIs, each with a short verbatim evidence span from
its description. Otherwise, the requirement is marked
\texttt{MISSING} and receives one query $q_i^{(j)}$
reformulating it for catalog resolution.
Requirements and queries are provider-neutral and contain
5--30 words; queries are not used as final functional roles.
An API may support multiple requirements.

Let $\mathcal{J}_i^{\mathrm{miss}}$ index the \texttt{MISSING}
records in $\mathcal{R}_i$. In checklist order, the
corresponding missing-item sequence is
\begin{equation}
  \mathcal{M}_i
  =
  \bigl(
    \rho_i^{(j)},
    e_i^{(j)},
    q_i^{(j)}
  \bigr)_{j\in\mathcal{J}_i^{\mathrm{miss}}}.
  \label{eq:missing-requirements}
\end{equation}
The Auditor returns \texttt{COMPLETE} when
$\mathcal{J}_i^{\mathrm{miss}}=\varnothing$; otherwise,
it returns \texttt{RESOLVE\_BATCH} with all items in
$\mathcal{M}_i$, without a fixed cardinality cap.
Before resolution, a deterministic validator checks the output
schema, sequential identifiers, evidence-substring constraints,
query lengths, and consistency between requirement statuses
and the overall decision. The full checklist and its supporting
evidence are recorded in $\tau_i$, while $\mathcal{M}_i$ is
passed to the Resolver.

\subsection{Catalog-Constrained Resolution}
\label{sec:method-resolver}

To constrain API proposals to the catalog while allowing
abstention, the Resolver represents each documented API $a$
by a dedicated action token $t_a$, such as
\texttt{<API\_google\_maps>}. For each missing requirement,
it selects an eligible API token or \texttt{<NO\_ADD>}
in the same action space.
It is trained to abstain when a separate API is unnecessary
or no eligible catalog API supports the queried capability
in the current recommendation state.

Let $\mathcal{T}=\{t_a:a\in\mathcal{A}\}$ denote the API action
tokens, and let $\mathcal{U}_i=\{t_a:a\in S_i^0\}\subseteq\mathcal{T}$
contain the tokens of all initial candidates, retained or deleted.
For each $j\in\mathcal{J}_i^{\mathrm{miss}}$, the Resolver input is
\begin{equation}
  x_i^{(j)}
  =
  \bigl(T_i,q_i^{(j)},e_i^{(j)},
  S_i^{\mathrm{keep}},\mathcal{U}_i\bigr).
  \label{eq:resolver-input}
\end{equation}
The input encodes $S_i^{\mathrm{keep}}$ and $\mathcal{U}_i$
as the initial API tokens, each labeled \texttt{KEEP} or
\texttt{DELETE} according to the Role Writer's decision.
All tokens in $\mathcal{U}_i$ are excluded from the active
action space
\begin{equation}
  \Omega_i
  =
  (\mathcal{T}\setminus\mathcal{U}_i)
  \cup\{\texttt{<NO\_ADD>}\}.
  \label{eq:resolver-space}
\end{equation}
At the first decoding step, let
$z_\theta(t\mid x_i^{(j)})$ denote the next-token
logit assigned to action $t$ by the Resolver.
We mask all tokens outside $\Omega_i$ and
select the highest-scoring eligible action:
\begin{equation}
  \widehat{t}_i^{(j)}
  =
  \arg\max_{t\in\Omega_i}
  z_\theta(t\mid x_i^{(j)}).
  \label{eq:resolver-selection}
\end{equation}

All missing queries are resolved in one batch against the
same fixed $S_i^{\mathrm{keep}}$ and $\mathcal{U}_i$.
We discard abstentions and deduplicate API identities
to form the candidate addition set
\begin{equation}
  \mathcal{C}_i
  =
  \bigl\{a\in\mathcal{A}:
    \exists j\in\mathcal{J}_i^{\mathrm{miss}},
    \ \widehat{t}_i^{(j)}=t_a
  \bigr\}.
  \label{eq:candidate-additions}
\end{equation}
Each candidate retains its originating requirement--query
associations for Commit Verification.

\emph{Resolver supervision.}
\label{sec:method-training}
Training examples combine reviewed requirement API pairs
from training mashups with short functional anchors for
documented catalog APIs; inputs follow
Eq.~\eqref{eq:resolver-input}. When multiple APIs independently
satisfy the same atomic requirement, their action tokens are
valid positives. Requirements needing complementary
contributions from multiple APIs are split into atomic
requirements.
For an example $x$, let $\Omega(x)$ be the active action
space and $\mathcal{P}(x)\subseteq\Omega(x)$ its nonempty
set of valid positive actions. We minimize the multi-positive loss
\begin{equation}
  \mathcal{L}_{\mathrm{res}}(x)
  =
  -\log
  \frac{
    \sum_{t\in\mathcal{P}(x)}
    \exp z_{\theta}(t\mid x)
  }{
    \sum_{t\in\Omega(x)}
    \exp z_{\theta}(t\mid x)
  }.
  \label{eq:resolver-loss}
\end{equation}
The loss maximizes total probability mass on valid actions
and reduces to cross-entropy over $\Omega(x)$ for a single positive.

\subsection{Commit Verification}
\label{sec:method-verifier}

The Resolver ensures catalog eligibility but neither reads
API descriptions at decision time nor generates functional
roles. A separate Commit Verifier therefore checks each proposal
against its catalog description before addition and writes
roles for accepted APIs. For each distinct $a\in\mathcal{C}_i$,
it receives $T_i$, $D_a$, and all originating queries in one
LLM call, with API names withheld. Queries specify the
capabilities to check; only $D_a$ serves as evidence of support.

The Verifier assesses each query separately, assigning
\texttt{SUPPORTED} if it judges that $D_a$ directly supports
the entire atomic capability in the context of $T_i$, and
\texttt{UNSUPPORTED} otherwise. Supported judgments include
evidence from $D_a$; unsupported ones include a brief rationale
based on $D_a$. We assign \texttt{ACCEPT} to an API if at least
one originating query is \texttt{SUPPORTED}, and
\texttt{REJECT} otherwise. Acceptance does not imply support
for every originating query. The accepted addition set is
\begin{equation}
  \Delta_i^+
  =
  \{a\in\mathcal{C}_i:
    a\text{ receives }\texttt{ACCEPT}\}.
  \label{eq:accepted-additions}
\end{equation}

For each accepted API, the Verifier writes $\eta_i(a)$ using
$T_i$ and $D_a$ to describe one atomic contribution from a
supported query--API match. When multiple queries are supported,
one representative contribution is selected for the role, while
all query-level judgments are retained in $\tau_i$.
The final recommendation is
$\widehat{S}_i=S_i^{\mathrm{keep}}\cup\Delta_i^+$,
with exactly one functional role for each API.

\emph{Deterministic controller.}
A deterministic controller applies the masking, deduplication, and acceptance
rules above and commits each accepted API once. It assembles
$\tau_i$ from the structured outputs of all stages, including
filtering decisions and deletion reasons, requirement checklist and
coverage evidence, the resolution decisions and verification outcomes. The controller
has no learned parameters and makes no semantic relevance
judgments.

\section{Experiments}
\label{sec:evaluation}

We evaluate \textsc{WAR-CRAFT} through
five research questions:

\noindent\textbf{RQ1}
Does \textsc{WAR-CRAFT} improve recommendation quality across
heterogeneous upstream recommenders, and can changes in set
size alone explain the gains?
(Section~\ref{sec:test-results})

\noindent\textbf{RQ2}
How does \textsc{WAR-CRAFT} compare with independent
refinement methods and controlled variants?
(Section~\ref{sec:exp-baselines})

\noindent\textbf{RQ3}
How do set-aware filtering, coverage-driven addition,
and commit verification affect recommendation quality?
(Section~\ref{sec:exp-components})

\noindent\textbf{RQ4}
Are the generated functional roles aligned with the request,
grounded in API documentation, and clearly expressed?
(Section~\ref{sec:exp-roles})

\noindent\textbf{RQ5}
How does initial set size affect refinement?
(Section~\ref{sec:exp-topk})

\begin{table*}[t]
  \centering
  \caption{
   Results on 1,643 test mashups.
  + \textsc{WAR-CRAFT} denotes the final outputs. Metrics are macro-averaged over mashups;
  Mean reports the average across seven sources.
  Cov. and Eff. denote functional Coverage and Efficiency,
  CovF1 their per-mashup harmonic mean, and Red. functional
  Redundancy ($\downarrow$).
  Bold marks the better quality score within each pair.
  }
  \label{tab:main-results}
  \small
  \setlength{\tabcolsep}{4pt}
  \begin{tabular}{llrrrrrrrrrr}
    \toprule
    Base & System & SetP & SetR & SetF1 & NDCG & MAP & Cov. & Eff.
      & CovF1 & Red.~$\downarrow$ & Avg.~$|S|$ \\
    \midrule
    WAR-R1 & Initial
      & 0.7282 & 0.6771 & 0.6800 & 0.7032 & 0.6643
      & 0.4414 & 0.5874 & 0.4574 & 0.0657 & 1.70 \\
      & + \textsc{WAR-CRAFT}
      & \textbf{0.7739} & \textbf{0.6937} & \textbf{0.7038}
      & \textbf{0.7145} & \textbf{0.6737}
      & \textbf{0.4439} & \textbf{0.6138} & \textbf{0.4718}
      & \textbf{0.0444} & 1.50 \\
    \midrule
    LLMAR & Initial
      & 0.7215 & 0.6734 & 0.6689 & 0.6909 & 0.6514
      & 0.4363 & 0.5542 & 0.4370 & 0.0700 & 1.95 \\
      & + \textsc{WAR-CRAFT}
      & \textbf{0.7476} & \textbf{0.6927} & \textbf{0.6911}
      & \textbf{0.7076} & \textbf{0.6661}
      & \textbf{0.4509} & \textbf{0.5830} & \textbf{0.4617}
      & \textbf{0.0690} & 1.75 \\
    \midrule
    GPT & Initial
      & 0.6544 & 0.5804 & 0.5912 & 0.5987 & 0.5641
      & 0.4085 & 0.5532 & 0.4309 & \textbf{0.0417} & 1.35 \\
      & + \textsc{WAR-CRAFT}
      & \textbf{0.6727} & \textbf{0.6095} & \textbf{0.6139}
      & \textbf{0.6254} & \textbf{0.5892}
      & \textbf{0.4320} & \textbf{0.5708} & \textbf{0.4497}
      & 0.0483 & 1.45 \\
    \midrule
    CVH-REC & Initial
      & 0.4537 & 0.5766 & 0.4698 & 0.5939 & 0.5479
      & \textbf{0.4731} & 0.4994 & 0.4564 & 0.1231 & 2.00 \\
      & + \textsc{WAR-CRAFT}
      & \textbf{0.6863} & \textbf{0.6430} & \textbf{0.6271}
      & \textbf{0.6553} & \textbf{0.6082}
      & 0.4577 & \textbf{0.5985} & \textbf{0.4788}
      & \textbf{0.0644} & 1.60 \\
    \midrule
    MARS & Initial
      & 0.5313 & 0.6724 & 0.5506 & 0.6970 & 0.6558
      & 0.4991 & 0.5490 & 0.4812 & 0.1912 & 2.00 \\
      & + \textsc{WAR-CRAFT}
      & \textbf{0.6423} & \textbf{0.6790} & \textbf{0.6194}
      & \textbf{0.6977} & \textbf{0.6559}
      & \textbf{0.4992} & \textbf{0.5917} & \textbf{0.4971}
      & \textbf{0.1150} & 1.81 \\
    \midrule
    WARBERT & Initial
      & 0.4169 & 0.5211 & 0.4274 & 0.5388 & 0.4965
      & 0.4385 & 0.4754 & 0.4271 & 0.1492 & 2.00 \\
      & + \textsc{WAR-CRAFT}
      & \textbf{0.6621} & \textbf{0.6285} & \textbf{0.6096}
      & \textbf{0.6350} & \textbf{0.5888}
      & \textbf{0.4529} & \textbf{0.5930} & \textbf{0.4715}
      & \textbf{0.0704} & 1.61 \\
    \midrule
    \toolft{} & Initial
      & 0.4860 & \textbf{0.6518} & 0.5184 & \textbf{0.6804}
      & \textbf{0.6318} & \textbf{0.4481} & 0.5087 & 0.4452 & 0.2446
      & 2.00 \\
      & + \textsc{WAR-CRAFT}
      & \textbf{0.7275} & 0.6309 & \textbf{0.6462} & 0.6557 & 0.6145
      & 0.4323 & \textbf{0.6247} & \textbf{0.4700}
      & \textbf{0.0466} & 1.40 \\
    \midrule
    Mean & Initial
      & 0.5703 & 0.6218 & 0.5580 & 0.6433 & 0.6017
      & 0.4493 & 0.5325 & 0.4479 & 0.1265 & 1.86 \\
      & + \textsc{WAR-CRAFT}
      & \textbf{0.7018} & \textbf{0.6539} & \textbf{0.6444}
      & \textbf{0.6702} & \textbf{0.6281}
      & \textbf{0.4527} & \textbf{0.5965} & \textbf{0.4715}
      & \textbf{0.0654} & 1.59 \\
    \bottomrule
  \end{tabular}
\end{table*}

\subsection{Experimental Setup}
\label{sec:exp-setup}

\emph{Dataset and split.}
We use the ProgrammableWeb corpus, the standard benchmark for Web API
recommendation~\cite{xu2026warbert,wu2021mashup,wu2025sefr}, with 8,217
mashups and 1,647 APIs. A mashup uses 2.09 recorded APIs on average, and mashup
and API descriptions average 17.0 and 44.3 words. We randomly split the mashups into 4,930 training,
1,644 validation, and 1,643 test instances (3:1:1),
using the same test set for all comparisons.

\emph{Upstream recommenders.}
We refine the frozen outputs of seven upstream recommenders.
The ranking sources are the hierarchical BERT-based recommender
WARBERT~\cite{xu2026warbert}, cross-view hypergraph recommender
CVH-REC~\cite{gao2026cvh}, multi-agent recommender
MARS~\cite{liu2026mars}, and \toolft{}, a semantic retriever
fine-tuned from Tool-Embed-4B~\cite{lu2026tools} on our training
split. The generative sources are LLMAR~\cite{qin2025harnessing},
WAR-R1~\cite{xu2026ranking}, and zero-shot GPT-5.6 Sol.
For ranking sources, the initial set contains the top-2 APIs,
close to the corpus average of 2.09; Section~\ref{sec:exp-topk}
varies this cutoff. For generative sources, we retain the
generated sets. At inference, \textsc{WAR-CRAFT} uses only
the predicted APIs and their order, mashup requests, and
catalog descriptions. 

\emph{Refinement baselines.}
All baselines use the same frozen upstream predictions
(Appendix~\ref{app:baseline-details}).
\emph{Independent refinement} uses none of our components.
Tuned-$k$ keeps the first $k$ upstream APIs.
MMR-Cut~\cite{10.1145/290941.291025} and DPP-Cut~\cite{chen2018fast}
select subsets for relevance and diversity from top-10 ranked
lists or generated sets, with tuned size caps or gain thresholds.
LLM-Select uses GPT-5.6 Sol to select from ranking sources'
top-10 APIs using the request and API descriptions, following
LLM-based reranking~\cite{sun2023chatgpt,gao2025llm4rerank}.
Retrieve-and-Select (R\&S) applies the same selector to the
initial set plus ten new catalog APIs retrieved for the request.
PTR-Refine adapts PTR~\cite{gao2025task} to the initial set,
omitting bundle acquisition. It retains APIs matched to
requirements and selects additions for unmatched requirements
through multi-view retrieval voting. \emph{Controlled variants.}
Direct-Edit replaces the Role Writer and Auditor with one
editing call; Self-Check adds a call to review its draft.
Separated extracts requirements from the request alone and
assesses coverage in a separate call.
Ret-Abstain replaces only the Resolver with a Tool-Embed-4B
retriever trained on the same supervision, scoring APIs
alongside a learned no-add candidate analogous to
Toolken+'s reject option~\cite{yakovlev2024toolken+}.
Other retrieval and embeddings use \toolft{}.
Independent baselines are tuned per source on validation SetF1
where applicable; \textsc{WAR-CRAFT} and its variants use
fixed settings across sources.

\emph{Resolver supervision.}
We construct supervision from recommendations on the original
mashup training split using CVH-REC and zero-shot GPT-5.6 Sol.
To broaden supervision beyond the evaluated upstreams, we
also include an auxiliary Qwen3-4B-Instruct-2507 recommender
trained following LLMAR~\cite{qin2025harnessing}. Only the first two also serve
as evaluated upstreams. CVH-REC and the Qwen recommender use
three-fold out-of-fold predictions: each mashup is predicted
by a model trained on the other two folds.
The Role Writer and Joint Requirement Auditor process these
recommendations. GPT-5.6 Sol reviews the generated queries in their
recommendation contexts, followed by author verification.
After review and deduplication, the retained supervision
comprises 3,434 API-action states (310 with multiple valid
targets) and 3,946 \texttt{<NO\_ADD>} states.
We add 436 catalog anchors, each pairing a standalone
capability query rewritten from an API description
with its API action token, yielding 7,816 states.
All anchors remain in training; 10\% of the reviewed
recommendation states are held out for Resolver
validation, separate from the mashup validation split
(Appendix~\ref{app:resolver-training}).

\emph{Implementation details.}
The Role Writer, Joint Requirement Auditor, and Commit
Verifier use GPT-5.6 Sol with reasoning effort set to none,
temperature 0. Their structured outputs are checked
against predefined schemas. We average recommendation
metrics over three runs of each GPT-based refinement method
on the same initial sets. The Resolver is fine-tuned from
Llama-3.2-3B-Instruct with LoRA; API action-token embeddings
are initialized from API names and descriptions.
We select the checkpoint with the highest state-level Hit@1
on held-out Resolver states and use it for all seven upstreams.
A hit matches any valid target action.

\emph{Metrics.}
Our primary metrics measure agreement between $\widehat{S}_i$
and the recorded API set $Y_i$: SetPrecision (SetP),
SetRecall (SetR), and SetF1, computed per mashup and
macro-averaged; we also report mean output size.
NDCG and MAP assess recovery and ranking over the complete
ordered output without a fixed cutoff. NDCG uses the ideal
DCG of the full recorded set, and average precision is
normalized by $|Y_i|$. Retained APIs keep their upstream
order, followed by accepted additions in audited
requirement order.

To assess functional alignment beyond recorded API membership,
we use DOEM's 53-label taxonomy~\cite{rong2026doem}, allowing
functional matches through APIs outside $Y_i$.
Coverage measures the fraction of required labels provided,
Efficiency the fraction of provided labels required, and
Redundancy the share of repeated label occurrences.
CovF1 is the harmonic mean of Coverage and Efficiency.
These metrics are computed per mashup and macro-averaged.

Functional roles are scored by a Qwen3-235B-A22B-Instruct-2507 judge, from a
different model family than GPT-5.6, to reduce
self-preference bias. The judge
sees only the mashup description, the API description, and the role,
without the system identity, and rates mashup alignment, API
grounding, and role clarity ~\cite{xu2026ranking} from 0 to 2; the Functional Role Score (FRS) is their
normalized sum.
Definitions and protocols are in
Appendices~\ref{app:metrics} and~\ref{app:role-evaluation}.

\subsection{Overall Effectiveness (RQ1)}
\label{sec:test-results}

Table~\ref{tab:main-results} evaluates the same
\textsc{WAR-CRAFT} configuration on seven frozen upstream
recommenders and 1,643 test mashups.

\emph{Recommendation accuracy.}
\textsc{WAR-CRAFT} improves SetF1 for all seven sources,
raising the mean from 0.5580 to 0.6444 (+8.64 percentage
points; 95\% paired-bootstrap CI: [8.01, 9.28]). All
source-specific intervals exclude zero
(Appendix~\ref{app:per-source}). Gains are larger for
the four ranking sources with top-2 inputs
(+6.88--18.22 points) than for the three generative
sources (+2.22--2.38 points).

\emph{Functional quality.}
Efficiency and CovF1 improve for all seven sources,
with mean CovF1 rising from 0.4479 to 0.4715.
Mean Redundancy falls from 0.1265 to 0.0654 and mean
output size from 1.86 to 1.59 APIs, although GPT's
Redundancy increases. For GPT, adding APIs with shared
functional labels can produce large ratio increases
in small sets (Appendix~\ref{app:per-source}). Mean Coverage changes little
(+0.34 percentage points),
with decreases for CVH-REC and \toolft{} despite
their CovF1 gains.

\emph{Set size and membership.}
Since 53.8\% of test mashups have one recorded API, we test whether prefix-based size adjustment explains
the gains (Table~\ref{tab:size}). Controls truncate upstream top-10 lists for ranking sources and original
generated lists otherwise to $|\widehat{S}_i|$ or
$|Y_i|$ (oracle size); shorter lists remain intact.
Within each source, exact matching to $|\widehat{S}_i|$
uses only instances with at least that many candidates.
On these instances (93.7\% on average across sources),
\textsc{WAR-CRAFT} exceeds equal-length prefixes by a mean
4.14 SetF1 percentage points
(95\% paired-bootstrap CI: [3.61, 4.67]).
For multi-API mashups ($|Y_i|\ge2$), the corresponding
eligible fraction and mean gain are 92.2\% and 4.22 points
(95\% CI: [3.54, 4.89]).
Across all multi-API mashups, \textsc{WAR-CRAFT} also
improves mean SetF1 and Coverage over Initial while
reducing mean output size from 2.09 to 1.93 APIs.
The gains therefore extend to multi-API mashups and persist
under exact size matching on eligible instances. \textsc{WAR-CRAFT} also exceeds truncation to the recorded set size in SetF1 (0.6444 versus 0.6240), although that oracle uses more APIs on multi-API mashups (2.71 versus 1.93).

\begin{table}[t]
\caption{Set size versus membership, averaged over seven sources.
Results include instances without enough candidates for exact
size matching. Cov., CovF1, and $|S|$ are on $|Y_i|\ge2$.
The oracle uses recorded sizes and is excluded from bolding.}
\label{tab:size}
\small
\setlength{\tabcolsep}{3pt}
\begin{tabular*}{\columnwidth}{@{\extracolsep{\fill}}lcccccc@{}}
\toprule
 & \multicolumn{3}{c}{SetF1} & \multicolumn{3}{c}{$|Y_i|\ge 2$}\\
\cmidrule(lr){2-4}\cmidrule(lr){5-7}
Method & All & $|Y_i|{=}1$ & $|Y_i|{\ge}2$ & Cov. & CovF1 & $|S|$\\
\midrule
Initial & 0.5580 & 0.5960 & 0.5137 & 0.4699 & 0.4574 & 2.09\\
Trunc.\ to $|\widehat{S}_i|$ & 0.6009 & 0.6767 & 0.5126 & 0.4423 & 0.4473 & 1.87\\
\textsc{WAR-CRAFT} & \textbf{0.6444} & \textbf{0.7206} & \textbf{0.5557}
  & \textbf{0.4835} & \textbf{0.4790} & 1.93\\
\midrule
Trunc.\ to $|Y_i|$ (oracle) & 0.6240 & 0.7105
  & 0.5233 & 0.5031 & 0.4559 & 2.71\\
\bottomrule 
\end{tabular*}
\end{table}


\begin{table}[t]
  \centering
  \caption{
  Refinement results averaged over seven upstream sources.
  Calls reports mean GPT-5.6 Sol calls per mashup during
  refinement. Bold marks the best quality scores.
  }
  \label{tab:baselines}

  \small
  \setlength{\tabcolsep}{3pt}
  \renewcommand{\arraystretch}{1.0}

  \begin{tabular*}{\columnwidth}
    {@{\extracolsep{\fill}}lrrrrr@{}}
    \toprule
    Method & SetF1 & CovF1 & Red.~$\downarrow$
           & Avg.~$|S|$ & Calls \\
    \midrule

    Initial
      & 0.5580 & 0.4479 & 0.1265 & 1.86 & 0 \\

    \midrule
    \multicolumn{6}{@{}l}{\emph{Independent refinement}} \\

    Tuned-$k$
      & 0.5982 & 0.4228 & \textbf{0.0227} & 1.23 & 0 \\
    MMR-Cut 
      & 0.6243 & 0.4349 & 0.0254 & 1.27 & 0 \\
    DPP-Cut
      & 0.6252 & 0.4365 & 0.0238 & 1.26 & 0 \\
    
    R\&S
      & 0.5334 & 0.4631 & 0.0924 & 1.63 & 1.00 \\
    PTR-Refine
      & 0.5625 & 0.4568 & 0.1118 & 1.76 & 2.71 \\

    \midrule
    \multicolumn{6}{@{}l}{\emph{Controlled variants}} \\

  Direct-Edit
      & 0.6092 & 0.4677 & 0.1115 & 1.72 & 1.22 \\
    + Self-Check
      & 0.6054 & 0.4621 & 0.1074 & 1.67 & 2.22 \\
    Separated
      & 0.6250 & 0.4704 & 0.0874 & 1.66 & 3.20 \\
Ret-Abstain
      & 0.6174 & 0.4582 & 0.0583 & 1.48 & 2.15 \\
    \midrule
    \textsc{WAR-CRAFT}
      & \textbf{0.6444} & \textbf{0.4715}
      & 0.0654 & 1.59 & 2.23 \\

    \bottomrule
  \end{tabular*}
\end{table}
\subsection{Refinement Baseline Comparison (RQ2)}
\label{sec:exp-baselines}
Table~\ref{tab:baselines} compares \textsc{WAR-CRAFT}
with five independent baselines and four controlled variants
across seven sources; per-source results are in
Appendix~\ref{app:per-source-baselines}.
\textsc{WAR-CRAFT} leads in mean SetF1 and CovF1,
exceeding DPP-Cut, the best independent baseline by SetF1,
by 1.92 percentage points (95\% CI: [1.26, 2.60]).

\emph{Independent refinement.}
On average, Tuned-$k$, MMR-Cut, and DPP-Cut reduce set
size and Redundancy but lower CovF1 relative to Initial.
DPP-Cut outperforms \textsc{WAR-CRAFT} on single-API
mashups (SetF1: 0.7355 versus 0.7206), but underperforms
Initial on multi-API mashups (0.4967 versus 0.5137).
LLM-Select also trails \textsc{WAR-CRAFT} on the four
ranking sources (SetF1: 0.5759 versus 0.6256).
R\&S and PTR-Refine improve CovF1 over Initial, but
R\&S lowers SetF1 and PTR-Refine gains 0.45 points.
PTR-Refine traces show that separate requirement matches
can retain overlapping APIs, while approximate matches
can suppress retrieval for unmet requirements.
Multi-view voting without abstention or a separate
documentation check can also admit unsupported additions.

\emph{Controlled variants.}
Self-Check does not improve Direct-Edit's mean SetF1.
Separated uses the most GPT calls (3.20 versus 2.23 per mashup for \textsc{WAR-CRAFT}).
Direct-Edit, Self-Check, and Separated all have
lower mean SetF1 and CovF1 and higher mean
Redundancy than \textsc{WAR-CRAFT}.
The traces illustrate differences in set auditing. 
On \emph{globebop}, Direct-Edit retains both Google Maps
and Google Street View Image for street-level views,
whereas the Role Writer keeps only Google Maps.
Direct-Edit also requests fewer missing capabilities
than the joint audit on six of seven sources,
with traces showing omitted functions or composite queries.
Separated can mistake application-side ranking for
an unmet external-service requirement, whereas the
joint audit identifies the missing external data.
With the same inputs, supervision, and action mask,
Ret-Abstain trails \textsc{WAR-CRAFT} on all seven sources,
with a mean SetF1 gap of 2.70 points
(95\% CI: [2.26, 3.16]).
The Verifier rejects 24.0\% of its proposals, versus
4.6\% for the Resolver.
The Resolver's accuracy advantage persists against
this retriever with learned abstention.

\subsection{Component Contributions (RQ3)}
\label{sec:exp-components}

Table~\ref{tab:stage-ablation} reports cumulative results
averaged over seven sources.

\emph{Set-aware filtering.}
The Role Writer raises mean SetF1 from 0.5580 to 0.5911 and reduces Redundancy from 0.1265 to 0.0485 (0.0520 excluding empty outputs), but lowers Coverage from 0.4493 to 0.4036 and CovF1 from 0.4479 to 0.4396. SetF1 gains are concentrated in the four ranking sources (+4.96 to +6.96 percentage points); for the generative sources, SetF1 changes by at most 0.26 points.

\emph{Coverage-driven addition.}
The Joint Requirement Auditor and Resolver then improve SetF1 for all seven sources, raising the mean to 0.6443. Mean Coverage returns to approximately its initial level (0.4532 versus 0.4493), while CovF1 rises to 0.4715. Redundancy increases to 0.0658 but remains below its initial level. Averaged over three runs, the Resolver returns \texttt{<NO\_ADD>} for 69.7\% of the queries it receives.

\emph{Commit verification.}
Averaged over three runs, the Commit Verifier rejects 4.6\% of the 2,889 deduplicated proposals per run. A blinded human review of 100 rejected and 100 accepted proposals from one run upholds 85 rejections and 92 acceptances (Appendix~\ref{app:verifier-review}). Mean SetF1 changes only from 0.6443 to 0.6444, with CovF1 unchanged at the reported precision (0.4715). The Verifier primarily provides documentation-based screening and functional roles for accepted additions, rather than a substantial SetF1 gain.

\emph{Human audit of requirements and API support.}
To assess requirement level support directly, two authors
annotated 140 random and 70 challenge source--mashup traces.
Requirements were first drafted from requests alone,
followed by API support scoring blind to upstream sources,
system checklists and decisions, and recorded-set membership
(Appendix~\ref{app:requirement-audit}).
On 139 assessable random traces, the share of reference
requirements fully supported by at least one API rises
from 55.2\% in the initial sets to 62.5\% in the final sets
(+7.28 percentage points; 95\% CI [4.24, 10.64]),
while the share of unsupported API instances falls
from 26.8\% to 7.8\%.
Manual comparison with the system checklists identifies
full matches for 82.8\% of these 261 reference requirements,
partial matches for 10.3\%, and omissions for 6.9\%.

\begin{table}[t]
  \centering
  \caption{
Cumulative refinement results averaged over seven
  upstream sources. Bold marks the best quality scores.
  }
  \label{tab:stage-ablation}

  \small
  \setlength{\tabcolsep}{4pt}
  \renewcommand{\arraystretch}{1.0}

  \begin{tabular*}{\columnwidth}
    {@{\extracolsep{\fill}}lrrrr@{}}
    \toprule
    Stage & SetF1~$\uparrow$ & CovF1~$\uparrow$
          & Red.~$\downarrow$ & Avg.~$|S|$ \\
    \midrule
    Initial
      & 0.5580 & 0.4479 & 0.1265 & 1.86 \\
    + Role Writer
      & 0.5911 & 0.4396 & \textbf{0.0485} & 1.35 \\
    + Audit
      & 0.6443 & \textbf{0.4715} & 0.0658 & 1.60 \\
    + Verifier
      & \textbf{0.6444} & \textbf{0.4715} & 0.0654 & 1.59 \\
    \bottomrule
  \end{tabular*}
\end{table}

\begin{table}[t]
  \centering
  \caption{Functional role quality after refinement.
Scores are normalized to $[0,1]$ and averaged over APIs
within each nonempty mashup, then over mashups.
Mean rows average the seven sources.
Empty counts mashups with no output APIs;
the zero-inclusive row assigns them zero scores.}
  \label{tab:role-quality}
  \small
  \setlength{\tabcolsep}{4pt}

  \begin{tabular}{@{}lrrrrr@{}}
    \toprule
    Source & \shortstack{Empty}
      & Alignment & Grounding & Clarity & FRS \\
    \midrule
    WAR-R1        & 83  & 0.9543 & 0.9349 & 0.9672 & 0.9521 \\
    LLMAR         & 39   & 0.9468 & 0.9210 & 0.9608 & 0.9429 \\
    GPT           & 115 & 0.9516 & 0.9223 & 0.9590 & 0.9443 \\
    CVH-REC       & 31  & 0.9320 & 0.9185 & 0.9539 & 0.9348 \\
    MARS          & 25  & 0.9334 & 0.9123 & 0.9549 & 0.9336 \\
    WARBERT       & 47  & 0.9279 & 0.9111 & 0.9498 & 0.9296 \\
    \toolft{} & 35  & 0.9696 & 0.9419 & 0.9650 & 0.9589 \\
    \midrule
    Mean                 & -- & 0.9451 & 0.9232 & 0.9587 & 0.9423 \\
    Mean, zero-inclusive & -- & 0.9142 & 0.8930 & 0.9274 & 0.9115 \\
    \bottomrule
  \end{tabular}
\end{table}

\subsection{Functional Role Quality (RQ4)}
\label{sec:exp-roles}

Table~\ref{tab:role-quality} reports final role quality
under the source-blind judge
(Appendix~\ref{app:role-evaluation}).
FRS ranges from 0.9296 to 0.9589 across sources.
Grounding scores lowest for all seven sources,
while clarity scores highest for six.
Assigning zero scores to empty outputs lowers
mean FRS from 0.9423 to 0.9115.
We evaluate WAR-R1's rationales~\cite{xu2026ranking}
with the same judge and scoring rubric, replacing only
``functional role'' with ``rationale'' in the prompt.
Its FRS is 0.9125 on recommended API instances with rationales, compared with 0.9521 for
\textsc{WAR-CRAFT}'s refined outputs from the same source.
However, 8.56\% of WAR-R1's recommended API instances lack rationales, while 7.53\% of API instances with
rationales are not recommended.
Assigning zero scores to these unmatched instances following \cite{xu2026ranking}
lowers its FRS to 0.8471.
By construction, every API in \textsc{WAR-CRAFT}'s
output carries exactly one functional role.

\emph{Human validation.}
Two authors independently rated 350 API--role instances
(50 per source), blinded to the source, writing component,
and judge scores, and resolved disagreements by discussion
(Appendix~\ref{app:human-validation}).
Across the three dimensions, the judge agrees exactly
with the consensus on 89.71--91.14\% of ratings,
with mean absolute errors of 0.089--0.109 on the 0--2 scale.
Quadratic-weighted Cohen's $\kappa$ ranges from
0.408 to 0.644, and Gwet's AC2 from 0.963 to 0.974. Consensus ratings are concentrated at the top of the scale:
89.90\% of dimension ratings receive the maximum score.

\subsection{Sensitivity to Initial Set Size (RQ5)}
\label{sec:exp-topk}
\begin{figure}[t]
  \centering
  \includegraphics[width=0.9\columnwidth]{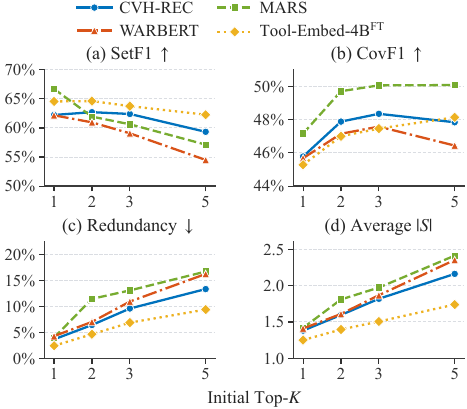}
  \vspace{-0.8em}
  \caption{
    Sensitivity to the initial cutoff $K$ under a fixed
    refinement configuration. (a) SetF1, (b) CovF1,
    (c) functional Redundancy, and (d) mean final set size.
  }
  \label{fig:topk}
\end{figure}

We vary the initial cutoff $K\in\{1,2,3,5\}$ for the four
ranking sources while keeping the refinement configuration
fixed. Figure~\ref{fig:topk} reports the final results,
averaged over three runs. SetF1 exceeds the corresponding
Initial in all 16 source--cutoff settings, by 1.25--30.56
percentage points. Although these gains increase with $K$,
final SetF1 peaks at $K=1$ or $2$ and declines by
2.28--9.57 points from $K=1$ to $K=5$. CovF1 exceeds
its $K=1$ value at every larger cutoff, gaining
0.76--2.92 points at $K=5$, while Redundancy and
final set size also increase. 
Stage analysis shows that more initial APIs remain after
filtering at larger $K$, while accepted additions decrease
by 46.2--75.2\% from $K=1$ to $K=5$ across sources
(Appendix~\ref{app:sens}). The larger final sets therefore
reflect more retained initial APIs despite fewer additions.
These results reveal a trade-off between historical-set
agreement, functional alignment, and output compactness.

\section{Discussion and Limitations}
\label{sec:discussion}

\textsc{WAR-CRAFT} performs one refinement pass without
re-auditing the final set. Composite queries may remain
unresolved if the Auditor fails to decompose them, as
the Resolver selects at most one API per query.
The semantic stages can be adapted to open-weight LLMs
for local deployment. Our experiments use the proprietary
GPT-5.6 Sol to prioritize the quality of semantic judgments
and functional roles.
Historical API sets and coarse
functional labels cannot fully assess valid alternatives,
motivating broader expert annotation of acceptable APIs
for each requirement.
The LLM stages use a proprietary model that may have seen
ProgrammableWeb during pretraining. In an exploratory check, it could not recover the functions of
long-tail APIs from their names alone, but neither this check nor
withholding API names rules out memorization.

\section{Conclusion}
\label{sec:conclusion}
We presented \textsc{WAR-CRAFT} for refining Web API
recommendations without modifying upstream models.
It filters unsupported or redundant APIs and jointly
audits external-service requirements and their coverage.
For each identified gap, the Resolver selects a catalog
API or abstains; proposed additions are checked against
API documentation. Each output API carries a request-specific
functional role, with refinement decisions recorded in an
audit trace. Experiments across seven heterogeneous
recommenders show higher SetF1 and CovF1 and lower average
functional redundancy. LLM-based evaluation and human
validation support the mashup alignment, API grounding, and clarity
of the generated roles.

\clearpage  
\begingroup
\sloppy
\bibliography{cite}
\endgroup
\appendix

\section{Experimental Details}
\label{app:evaluation}

\subsection{Models and Training}
\label{app:upstream}
\label{app:resolver-training}
\emph{Upstream recommenders.}
Trained recommenders, including WAR-R1, use each method's released code
and default settings, with training on our training split and model
selection on our validation split; MARS keeps its default
Qwen2.5-14B-Instruct agents, and \toolft{} is Tool-Embed-4B further
fine-tuned with LoRA (rank 32) for five epochs. Zero-shot GPT-5.6 Sol
returns API names only. Names are normalized and matched to the catalog; for an unmatched
name, GPT-5.6 Sol states the intended function, which is mapped to a
catalog API or dropped if no API provides it. All upstream
outputs are generated once and frozen. Empty initial sets (40 for
LLMAR, 5 for WAR-R1, and 169 for GPT) skip the Role Writer.

\emph{Supervision construction.}
For the training states in Section~\ref{sec:exp-setup}, GPT-5.6 Sol
reviews each Auditor query against the mashup, current and recorded
API sets, and catalog descriptions. Valid targets are recorded APIs
outside the initial set that independently satisfy the query;
multiple targets are allowed. 
A query receives \texttt{<NO\_ADD>} if its capability is
already covered, requires no separate API, or lacks
full documented support from any eligible catalog API. Queries that
only non-recorded APIs satisfy therefore receive neither label and are
excluded (36 states), as are other unresolved ambiguous examples. Composite
requirements are split when independently resolvable, and exact
duplicate inputs are merged. The 436 catalog anchors
(Section~\ref{sec:exp-setup}) cover APIs absent from the reviewed
targets, including 143 appearing in recorded test sets. All anchors
remain in training; the 10\% validation subset contains only reviewed
recommendation states.

\emph{Training details.}
We fine-tune the Resolver from Llama-3.2-3B-Instruct using LoRA with
rank 32, $\alpha=64$, dropout 0.05, and a learning rate of
$5\times10^{-5}$. New API action-token embeddings are jointly updated.
Training runs for four epochs on two RTX 4090 GPUs
with seed 42, taking approximately 21 minutes.
Token initialization and checkpoint selection follow
Section~\ref{sec:exp-setup}.

\emph{Computational cost.}
The Resolver averages 0.04\,s of inference time per
mashup. Mean input/output token counts per GPT call
are 1,215/67 for the Role Writer, 1,059/210 for the
Joint Requirement Auditor, and 924/76 for the
Commit Verifier.

\subsection{Refinement Baseline Details}
\label{app:baseline-details}

LLM-based baselines use GPT-5.6 Sol with the decoding settings of Section~\ref{sec:exp-setup} and are run three times; retrieval and embeddings use \toolft{}. Hyperparameters are selected per source by validation macro SetF1; Table~\ref{tab:refinement-selected-settings} lists the selected values.

\begin{table}[t]
  \centering
  \caption{ Validation selected settings. Entries are $(\gamma,w, \delta)$, with $w=\lambda$ for MMR and $w=\theta$ for DPP; $\gamma=0/1$ denotes rank/semantic relevance. GPT's DPP uses a cap of 10 instead of a threshold.}
  \label{tab:refinement-selected-settings}
  \small
  \setlength{\tabcolsep}{3pt}
\begin{tabular*}{\columnwidth}
  {@{\extracolsep{\fill}}lccc@{}}
  \toprule
  Source & Tuned-$k$ & MMR-Cut & DPP-Cut \\
  \midrule
  LLMAR
    & 4 & $(1,0.75,0.225)$ & $(1,0.90,0.400)$ \\
  CVH-REC
    & 1 & $(1,0.50,-0.075)$ & $(1,0.90,0.475)$ \\
  GPT
    & 4 & $(1,1,0.550)$ & $(0,0.10,K{=}10)$ \\
  MARS
    & 1 & $(0,0.25,-0.425)$ & $(0,0.50,-0.125)$ \\
  WARBERT
    & 1 & $(1,0.50,-0.075)$ & $(1,0.90,0.475)$ \\
  \toolft{}
    & 1 & $(0,0.50,-0.125)$ & $(0,0.50,-0.350)$ \\
  WAR-R1
    & 3 & $(1,0.75,0.250)$ & $(1,0.90,0.400)$ \\
  \bottomrule
\end{tabular*}
\end{table}

\textbf{1) \emph{Tuned-$k$, MMR-Cut, and DPP-Cut.}}
Tuned-$k$ retains up to $k$ APIs in upstream output order, $k\in\{1,2,3,4\}$. MMR-Cut and DPP-Cut use the upstream top-10 ranked list or original generated set, without additional retrieval. For each source and algorithm, validation selects either semantic relevance $u_a=(1+\cos(e_T,e_a))/2$ or discounted rank $u_a=1/\log_2(1+r_a)$. API similarity is a shifted cosine Gram matrix regularized toward the identity with weight $10^{-6}$. MMR greedily trades relevance, weighted by $\lambda\in\{0.1,0.25,0.5,0.75,0.9,1\}$, against maximum similarity to selected APIs; DPP greedily maximizes the log-determinant of a kernel combining relevance, weighted by $\theta\in\{0.1,0.25,0.5,0.75,0.9,0.95\}$, with API similarity. Both stop at a size cap in $\{1,2,\ldots,10\}$, bounded by the available candidates, or when marginal gain falls below $\delta\in[-0.5,1.0]$ (step 0.025).
 \textbf{2) \emph{LLM-Select.}}
For each ranking recommender, the selector receives the request and the upstream's top-10 APIs with their catalog descriptions in random order, and returns a supported, complementary subset, possibly empty, in the order it chooses.
 \textbf{3) \emph{Retrieve-and-Select.}}
Retrieve-and-Select applies the LLM-Select procedure to $S_i^0\cup\mathrm{Top}_{B}(T_i;\mathcal{A}\setminus S_i^0)$, with $B=10$ distinct new APIs and initial-set membership hidden. Validation Recall@10 is 0.3713 for BM25, 0.5810 for released Tool-Embed-4B, and 0.7423 for \toolft{}, which is therefore used for all retrieval and embedding baselines.
 \textbf{4) \emph{PTR-Refine.}}
PTR-Refine adapts PTR~\cite{gao2025task} to the same initial sets, omitting historical-bundle acquisition. One prompt extracts requirements from the request, and one per requirement identifies initial APIs that roughly fulfill it. Only APIs matched to at least one requirement are retained. Each unmatched requirement drives three catalog retrieval views: requirement-to-API similarity, APIs of similar training mashups, and API-to-API similarity from the top direct hit. The API receiving the most votes is added unless already present, with ties broken by direct similarity. Validation selects retrieval depth $K=10$ from $\{3,5,10\}$.
 \textbf{5) \emph{Direct-Edit and Self-Check.}}
Direct-Edit receives the request and initial APIs with their descriptions, without access to the rest of the catalog. One prompt jointly produces KEEP/DELETE decisions, roles for retained APIs, and queries for missing capabilities. These queries pass through the same Resolver and Commit Verifier as in \textsc{WAR-CRAFT}. Self-Check adds one prompt that reviews the input and Direct-Edit draft and returns a corrected draft before resolution.
 \textbf{6) \emph{Separated.}}
Separated replaces the Joint Requirement Auditor with two prompts: requirement extraction from the request alone, followed by coverage assessment against the APIs retained by the Role Writer. 
\textbf{7) \emph{Ret-Abstain.}}
Ret-Abstain replaces only the Resolver. A shared encoder maps the input in Eq.~\eqref{eq:resolver-input} and each API profile to normalized vectors $q$ and $d_a$, respectively. With a fixed temperature $T=0.03$, each available API scores $s_a=q^\top d_a/T$, and a learned no-add candidate scores $s_{\varnothing}=q^\top r/(T\lVert r\rVert)+b$. It is trained for five epochs with cross-entropy on the Resolver's supervision states, updating LoRA adapters on Tool-Embed-4B (rank 16, $\alpha=32$, learning rate $10^{-5}$) and $r,b$ (learning rate $5\times10^{-3}$); API embeddings are refreshed every epoch, and the checkpoint with the best validation Hit@1 is kept. Ret-Abstain outputs \texttt{<NO\_ADD>} when $s_{\varnothing}\ge\max_a s_a$ and otherwise proposes the top API, which is merged and verified as in \textsc{WAR-CRAFT}.

\subsection{Metrics and Functional Annotations}
\label{app:metrics}

\emph{Historical-set agreement.}
For mashup $i$, let $Y_i$ be the recorded API set and $\widehat{S}_i$ the predicted set. SetPrecision and SetRecall normalize $|\widehat{S}_i\cap Y_i|$ by $|\widehat{S}_i|$ and $|Y_i|$, respectively, and $\operatorname{SetF1}_i=2|\widehat{S}_i\cap Y_i|/(|\widehat{S}_i|+|Y_i|)$ is their harmonic mean.
All scores, including SetF1, are computed per mashup and macro-averaged.
Empty predictions receive zero for the set and ranking metrics.

\emph{Ranking metrics.}
\label{app:dynamic-ranking-metrics}
For the ordered output $(a_{i1},\ldots,a_{in_i})$ with $n_i=|\widehat{S}_i|$, retained APIs keep their upstream order and accepted additions follow in audited requirement order. With $g_{ik}=\mathbf{1}[a_{ik}\in Y_i]$, we compute $\operatorname{NDCG}_i=\operatorname{DCG}_i/\operatorname{IDCG}_i$, where $\operatorname{DCG}_i=\sum_{k=1}^{n_i}g_{ik}/\log_2(k+1)$ and $\operatorname{IDCG}_i=\sum_{k=1}^{|Y_i|}1/\log_2(k+1)$, and $\operatorname{AP}_i=\sum_{k=1}^{n_i}g_{ik}\,\mathrm{P}_i@k/|Y_i|$, where $\mathrm{P}_i@k$ is the precision of the first $k$ outputs. Both use the complete output without a fixed cutoff and normalize by all $|Y_i|$ recorded APIs. NDCG and MAP macro-average $\operatorname{NDCG}_i$ and $\operatorname{AP}_i$ across mashups.

\emph{Functional annotation.}
We use the fixed 53-label taxonomy from DOEM~\cite{rong2026doem}.
Of the 1,647 documented APIs, 1,151 inherit labels through exact
normalized name matching, and 496 are reviewed from their catalog
descriptions. A mashup inherits labels when its normalized name exactly matches a
single DOEM mashup; unmatched mashups are annotated from their
descriptions without access to historical API sets.

\emph{Functional metrics.}
Let $\mathcal{Q}_i$ be the mashup's required labels and $\mathcal{G}(a)$ the labels of API $a$, with $\mathcal{G}(\widehat{S}_i)=\bigcup_{a\in\widehat{S}_i}\mathcal{G}(a)$ and $\mathcal{H}_i=\mathcal{Q}_i\cap\mathcal{G}(\widehat{S}_i)$. Coverage is $\operatorname{Cov}_i=|\mathcal{H}_i|/|\mathcal{Q}_i|$, Efficiency is $\operatorname{Eff}_i=|\mathcal{H}_i|/|\mathcal{G}(\widehat{S}_i)|$, and $\operatorname{CovF1}_i$ is their harmonic mean, set to 0 when both are 0. Redundancy is $\operatorname{Red}_i=1-|\mathcal{G}(\widehat{S}_i)|/\sum_{a\in\widehat{S}_i}|\mathcal{G}(a)|$, the share of repeated label occurrences. We set $\operatorname{Eff}_i=0$ for an empty provided-label set and $\operatorname{Red}_i=0$ for an empty recommendation set, and compute all four metrics per mashup before macro-averaging.

\section{Additional Results}
\label{app:per-source}

\emph{Overall effectiveness.}
Table~\ref{tab:per-source-ci} reports gains over Initial.
For the primary comparisons, we first average
per-mashup metric values across three runs for each
GPT-based refinement method. We compute 95\% paired
bootstrap confidence intervals using 20,000
mashup-level resamples (seed 2026). Each resample
uses the same mashup IDs across methods and upstream
sources. For the Mean row, we average the seven source-specific
gains within each resample. Mean Coverage shows
no significant change.
Coverage decreases for CVH-REC and \toolft{}, while
Redundancy increases only for GPT. A paired analysis
of GPT outputs finds increases in 55 of 1,643 mashups
and decreases in 25. The increasing cases receive
one (43 cases) or two (12 cases) additional APIs.
Their mean Redundancy increase is 0.2502
(median 0.2500; interquartile range 0.1667--0.3333).
Among these cases, 28 initially contain a single API
and account for 57.2\% of the summed positive changes,
illustrating how a few additions sharing functional
labels can sharply raise the ratio in GPT's small
initial sets.

\begin{table*}[t]
  \centering
  \caption{ Per-source gains of \textsc{WAR-CRAFT} over Initial
  in percentage points, with 95\% paired-bootstrap
  confidence intervals. Negative $\Delta$Red.
  indicates reduced functional redundancy.}
  \label{tab:per-source-ci}
  \small
\begin{tabular}{lcccc}
  \toprule
  Source & $\Delta$SetF1 & $\Delta$CovF1 & $\Delta$Cov. & $\Delta$Red. \\
  \midrule
  WAR-R1
    & $+2.38$ $[1.52, 3.24]$
    & $+1.45$ $[0.84, 2.06]$
    & $+0.25$ $[-0.50, 1.00]$
    & $-2.13$ $[-2.77, -1.51]$ \\
  LLMAR
    & $+2.22$ $[1.43, 3.02]$
    & $+2.47$ $[1.91, 3.03]$
    & $+1.46$ $[0.81, 2.12]$
    & $-0.10$ $[-0.54, 0.35]$ \\
  GPT
    & $+2.27$ $[1.58, 3.01]$
    & $+1.87$ $[1.40, 2.38]$
    & $+2.35$ $[1.83, 2.88]$
    & $+0.66$ $[0.40, 0.93]$ \\
  CVH-REC
    & $+15.74$ $[14.43, 17.01]$
    & $+2.24$ $[1.36, 3.13]$
    & $-1.54$ $[-2.54, -0.55]$
    & $-5.87$ $[-6.72, -5.01]$ \\
  MARS
    & $+6.88$ $[5.89, 7.84]$
    & $+1.59$ $[0.97, 2.20]$
    & $+0.01$ $[-0.70, 0.73]$
    & $-7.62$ $[-8.51, -6.76]$ \\
  WARBERT
    & $+18.22$ $[16.80, 19.65]$
    & $+4.44$ $[3.47, 5.44]$
    & $+1.44$ $[0.31, 2.60]$
    & $-7.88$ $[-8.80, -6.98]$ \\
  \toolft{}
    & $+12.79$ $[11.49, 14.11]$
    & $+2.48$ $[1.60, 3.37]$
    & $-1.59$ $[-2.60, -0.59]$
    & $-19.80$ $[-20.72, -18.87]$ \\
  \midrule
  Mean
    & $+8.64$ $[8.01, 9.28]$
    & $+2.36$ $[1.97, 2.75]$
    & $+0.34$ $[-0.12, 0.80]$
    & $-6.11$ $[-6.49, -5.71]$ \\
  \bottomrule
\end{tabular}
\end{table*}

\emph{Refinement baselines.}
\label{app:per-source-baselines}
Table~\ref{tab:per-source-baselines} reports SetF1 and
CovF1 by upstream source. \textsc{WAR-CRAFT} achieves
the highest SetF1 on all four ranking sources and
LLMAR, while Separated leads on WAR-R1 and GPT. For CovF1, \textsc{WAR-CRAFT} leads on five sources;
Retrieve-and-Select and Separated lead on GPT and
CVH-REC, respectively.

\begin{table*}[t]
  \centering
  \caption{
     Per-source SetF1 and CovF1 of refinement methods.
  Tool denotes \toolft{}, and R\&S denotes
  Retrieve-and-Select. Bold marks the highest value
  in each column; -- indicates not applicable.
  }
  \label{tab:per-source-baselines}

  \footnotesize
  \setlength{\tabcolsep}{2pt}
  \begin{tabular*}{\textwidth}
    {@{\extracolsep{\fill}}l*{14}{c}@{}}
    \toprule
    & \multicolumn{7}{c}{SetF1}
    & \multicolumn{7}{c}{CovF1} \\
    \cmidrule(lr){2-8}
    \cmidrule(lr){9-15}

    Method
      & WAR-R1 & LLMAR & GPT & CVH-REC & MARS & WARBERT & Tool
      & WAR-R1 & LLMAR & GPT & CVH-REC & MARS & WARBERT & Tool \\
    \midrule

    Initial
      & 0.6800 & 0.6689 & 0.5912 & 0.4698
      & 0.5506 & 0.4274 & 0.5184
      & 0.4574 & 0.4370 & 0.4309 & 0.4564
      & 0.4812 & 0.4271 & 0.4452 \\
    \midrule

    \multicolumn{15}{@{}l}{\emph{Independent refinement}} \\

    Tuned-$k$
      & 0.6802 & 0.6689 & 0.5907 & 0.5225
      & 0.6189 & 0.4737 & 0.6330
      & 0.4570 & 0.4388 & 0.4314 & 0.4055
      & 0.4277 & 0.3780 & 0.4214 \\

    MMR-Cut
      & 0.6809 & 0.6702 & 0.5912 & 0.6077
      & 0.6182 & 0.5618 & 0.6404
      & 0.4507 & 0.4400 & 0.4302 & 0.4196
      & 0.4492 & 0.4053 & 0.4493 \\

    DPP-Cut
      & 0.6810 & 0.6659 & 0.5912 & 0.6109
      & 0.6192 & 0.5646 & 0.6437
      & 0.4580 & 0.4366 & 0.4309 & 0.4301
      & 0.4438 & 0.4134 & 0.4429 \\

    LLM-Select
      & -- & -- & -- & 0.5875
      & 0.5847 & 0.5565 & 0.5750
      & -- & -- & -- & 0.4400
      & 0.4762 & 0.3987 & 0.4547 \\

   R\&S
      & 0.5647 & 0.5397 & 0.5242 & 0.5300
      & 0.5279 & 0.5292 & 0.5183
      & 0.4692 & 0.4616 & \textbf{0.4665}
      & 0.4615 & 0.4680 & 0.4609 & 0.4537 \\

    PTR-Refine
      & 0.6502 & 0.6409 & 0.6058 & 0.5060
      & 0.5441 & 0.4823 & 0.5084
      & 0.4617 & 0.4481 & 0.4550 & 0.4678
      & 0.4748 & 0.4442 & 0.4463 \\

    \midrule
    \multicolumn{15}{@{}l}{\emph{Controlled variants}} \\

    Direct-Edit
      & 0.6901 & 0.6794 & 0.6129 & 0.5905
      & 0.5763 & 0.5644 & 0.5509
      & 0.4716 & 0.4559 & 0.4531
      & 0.4809 & 0.4913 & 0.4683 & 0.4529 \\

    + Self-Check
      & 0.6906 & 0.6753 & 0.6004 & 0.5926
      & 0.5714 & 0.5550 & 0.5525
      & 0.4705 & 0.4559 & 0.4462 & 0.4721
      & 0.4820 & 0.4542 & 0.4536 \\

    Separated
      & \textbf{0.7065} & 0.6879
      & \textbf{0.6242}
      & 0.5935 & 0.5880 & 0.5734 & 0.6017
      & 0.4712 & 0.4613 & 0.4600
      & \textbf{0.4814}
      & 0.4946 & 0.4626 & 0.4618 \\
Ret-Abstain
      & 0.6953 & 0.6801 & 0.5969 & 0.5775
      & 0.6127 & 0.5715 & 0.5879
      & 0.4675 & 0.4557 & 0.4431
      & 0.4605 & 0.4916 & 0.4493 & 0.4397 \\
    \midrule

    \textsc{WAR-CRAFT}
      & 0.7038 & \textbf{0.6911} & 0.6139
      & \textbf{0.6271} & \textbf{0.6194}
      & \textbf{0.6096} & \textbf{0.6462}
      & \textbf{0.4718} & \textbf{0.4617} & 0.4497 & 0.4788
      & \textbf{0.4971} & \textbf{0.4715}
      & \textbf{0.4700} \\

    \bottomrule
  \end{tabular*}
\end{table*}

\emph{Initial set size sensitivity.}
\label{app:sens}
For $K=(1,2,3,5)$, SetF1 gains over the corresponding
initial sets, in percentage points, are
(9.98, 15.74, 21.41, 26.90) for CVH-REC,
(4.81, 6.88, 13.39, 20.99) for MARS,
(14.80, 18.22, 22.42, 25.68) for WARBERT, and
(1.25, 12.79, 21.08, 30.56) for \toolft{}.
Across sources, mean retained size increases from
0.775--0.940 APIs at $K=1$ to 1.532--2.271 at $K=5$.
Over the same range, missing-query totals decrease by
18.6--47.5\% and accepted API additions by 46.2--75.2\%.


\emph{Additional controls.}
We tested two common retrieval replacements that take \toolft{}'s top API for each missing query: always adding it lowers SetF1 to 0.5807, below filtering alone (0.5911), and adding it only when its cosine similarity reaches a per-source validation threshold (0.35 for every source) yields 0.5937. Ret-Abstain (0.6174) is the strongest retrieval-based replacement we tested.
We also tested reranking final sets containing at least
two APIs with GPT-5.6 Sol. Mean NDCG and MAP changed only
by -0.5 and -0.6 points, respectively. One possible explanation is that filtering removes
poorly matched candidates while preserving useful
upstream ordering, leaving limited benefit from
further reranking.

\section{Human Audit and Role Evaluation}

\subsection{Functional Role Evaluation}
\label{app:role-evaluation}

\emph{Scoring protocol.}
The source-blind Qwen3-235B-A22B-Instruct-2507 judge receives only the
mashup description, API description, and role. Using the two descriptions
as evidence, it evaluates the role without rewriting and independently
scores mashup alignment, API grounding, and clarity in $\{0,1,2\}$,
adapting WAR-R1's criteria~\cite{xu2026ranking}. Dimension scores are
divided by two; FRS is the sum of the three raw scores divided by six.
Scores are averaged over APIs within each mashup, then over mashups.
A fixed prompt
and temperature 0 are used across systems.

\emph{Human validation of the role Judge.}
\label{app:human-validation}
From one scored run's final outputs, we randomly sample 50 API--role instances per source: 350 from 316 distinct mashups. Two authors independently score them using the same inputs and rubric, blind to the source, role-writing component, and judge scores. Pre-discussion exact agreement on alignment, grounding, and clarity is 91.43\%, 92.00\%, and 92.86\%, respectively (quadratic-weighted $\kappa$ 0.490, 0.733, and 0.579; AC2 0.972, 0.974, and 0.979). The 83 disagreements among 1,050 ratings, spanning 64/350 instances, are resolved by discussion; Section~\ref{sec:exp-roles} compares the consensus with the judge.

\emph{Negative control.}
For each of the 350 human-validated instances $(T_i, D_a, r)$, we
keep $T_i,D_a$ and uniformly sample a replacement role $r'$ (seed 42)
from the same source's final outputs on other mashups. Donor API $b$
must satisfy $\mathcal{G}(b)\cap(\mathcal{G}(a)\cup\mathcal{Q}_i)=\emptyset$,
and $r'$ must differ from $r$; every instance has an eligible donor.
We reuse original scores and evaluate replacements with the same prompt
and judge settings. Mean normalized FRS falls
from 0.9486 to 0.2395, a drop of 0.7090 (95\% CI [0.6890, 0.7281];
10,000 paired bootstrap resamples, seed 42); FRS decreases for 343 of
the 350 instances (98.0\%) and is unchanged for the other seven.
Alignment and grounding fall to 0.0271 and 0.0100, whereas clarity
falls only to 0.6814, consistent with fluent but mismatched roles.

\subsection{Requirements and Refinement Decisions}
\label{app:requirement-audit}

We sampled 20 random and 10 challenge
traces per source (seed 2026), yielding 140 random and 70
challenge pairs covering 202 distinct mashups. Challenge
selection used redundancy, filtering, and unmet-requirement
signals, without recorded-set membership or performance scores.
Two authors each annotated 130 traces, with 50 independently
double-annotated. Requirements were drafted from requests alone
before API scoring; later additions or scope revisions were
logged for two random and three challenge entries. Shuffled,
anonymized API descriptions concealed sources, ranks, system
checklists and decisions, and recorded API sets.

API--requirement support was scored as absent (0), partial (1),
or full (2). Full coverage requires one API scoring 2 for the
entire reference entry, including compound entries; an unsupported
API scores 0 against every entry. Excluding one unconfirmed empty
reference leaves 139/70 assessable traces with 261/160 requirements
(random/challenge). Final statistics include all stored APIs.
An author assessed checklist alignment: full matches
capture all actions, objects, and key constraints of
a reference requirement; partial matches capture
only part of this scope; omissions have no
corresponding checklist content. Pre-discussion agreement
on 383 comparable support scores was 353/383 (92.2\%;
quadratic-weighted $\kappa=0.921$); disagreements were resolved
through discussion. Proportions pool counts within each stratum.
The coverage-gain CI uses 10,000 paired percentile-bootstrap
resamples clustered by mashup (seed 42).

Among scope-comparable decisions, false \texttt{COVERED} judgments
occurred in 8/90 random
and 8/54 challenge cases; false \texttt{MISSING} judgments
occurred in 13/74 and 7/48, respectively.
Filtering lost full coverage of 3/144 random requirements
(all recovered by later additions) and 1/84 challenge requirements
(not recovered). Eligible alternatives were confirmed for 10/37
random and 12/30 challenge genuine gap abstentions. The remaining
genuine-gap abstentions remain unclassified: review covered
trajectory APIs and generally the retrieved top three, expanded
to ten for 28 queries selected before scoring.

\emph{Review of verification decisions.}
\label{app:verifier-review}
We randomly sample and shuffle 100 rejected and 100 accepted proposals
from one run. Two annotators independently apply the Verifier's
query-level criterion using only the request, API description, and
originating queries, blind to API names, Verifier outputs, sources,
and recorded-set membership. A proposal is supported if at least one
query is supported. Pre-discussion agreement is 93.5\% (Cohen's
$\kappa=0.87$); the 13 disagreements are resolved by discussion,
and reported counts use consensus labels. Of 33 sampled rejections
of recorded APIs, 24 are upheld.

\end{document}